\documentclass[letterpaper]{article} 
\usepackage{aaai2026}  
\usepackage{times}  
\usepackage{helvet}  
\usepackage{courier}  
\usepackage[hyphens]{url}  
\usepackage{graphicx} 
\usepackage{natbib}  
\usepackage{caption} 
\usepackage{algorithm}
\usepackage{algorithmic}
\usepackage{amsmath}
\usepackage{color}
\usepackage{multirow}
\usepackage{booktabs} 
\usepackage{amssymb}
\usepackage{xcolor}         
\usepackage{newfloat}
\usepackage{listings}
\DeclareCaptionStyle{ruled}{labelfont=normalfont,labelsep=colon,strut=off} 
\floatstyle{ruled}
\newfloat{listing}{tb}{lst}{}
\floatname{listing}{Listing}
\title{\title{\texttt{Dream-IF}: Dynamic Relative EnhAnceMent for Image Fusion}}
\author{Xingxin Xu$^{1}$ \quad \textbf{Bing Cao}$^2$\thanks{Corresponding author.} \quad  \quad \textbf{DongDong Li}$^3$  \quad \textbf{Qinghua Hu}$^{2}$ \quad \textbf{Pengfei Zhu}$^{2,3,4}$\footnotemark[1] 
 \quad
}
\affiliations{
    \textsuperscript{\rm 1}School of New Media and Communication, Tianjin University, Tianjin, China\\
 \textsuperscript{\rm 2} School of Artificial Intelligence, Tianjin University\\
  \textsuperscript{\rm 3} Low-Altitude Intelligence Laboratory, Xiong'an National Innovation Center  \\
   \textsuperscript{\rm 4} Xiong'an Guochuang Lantian Technology Co., Ltd. \\
 \textsuperscript{\rm 5} National Key Laboratory of Science and Technology on Automatic Target Recognition, National University of Defense Technology, Changsha, China\\
   \texttt{\{xuxingxin, caobing, huqinghua, zhupengfei\}@tju.edu.cn, lidongdong12@nudt.edu.cn}
}

\usepackage{bibentry}

\begin{document}

\maketitle

\begin{abstract}

Image fusion aims to integrate comprehensive information from images acquired through multiple sources. However, images captured by diverse sensors often encounter various degradations that can negatively affect fusion quality. Traditional fusion methods generally treat image enhancement and fusion as separate processes, overlooking the inherent correlation between them; notably, the dominant regions in one modality of a fused image often indicate areas where the other modality might benefit from enhancement. Inspired by this observation, we introduce the concept of dominant regions for image enhancement and present a Dynamic Relative EnhAnceMent framework for Image Fusion (Dream-IF). This framework quantifies the relative dominance of each modality across different layers and leverages this information to facilitate reciprocal cross-modal enhancement. By integrating the relative dominance derived from image fusion, our approach supports not only image restoration but also a broader range of image enhancement applications. Furthermore, we employ prompt-based encoding to capture degradation-specific details, which dynamically steer the restoration process and promote coordinated enhancement in both multi-modal image fusion and image enhancement scenarios. Extensive experimental results demonstrate that Dream-IF consistently outperforms its counterparts.  The code is publicly available.\footnote{ \url{https://github.com/jehovahxu/Dream-IF}}

\end{abstract}
\section{Introduction}
\label{sec:intro}
Image fusion aims to integrate essential information from multi-source images captured by various sensors, producing a single comprehensive image. 
Multi-modal (visible \textit{v.s.} infrared) image fusion~\cite{zhao2023cddfuse, ma2022swinfusion, li2018densefuse} has been used in a wide range of applications, such as auto driving~\cite{huang2022multi}, unmanned aerial vehicles~\cite{jasiunas2002image}, forest fire monitoring~\cite{liu2023forest}, etc. Images captured by different imaging sensors often have different characteristics.
The infrared images intrinsically avoid visible-light interference but often lack fine texture details. In contrast, the visible images capture abundant color and details in well-lit areas, while may suffer significant quality degradation in complex environments. 
The fused image preserves the benefits of visible and infrared images while minimizing their limitations by utilizing the complementary information from the respective modalities~\cite{ma2019infrared,liu2024infrared}.
\begin{figure}[!t]
\centering
\includegraphics[width=1\linewidth]{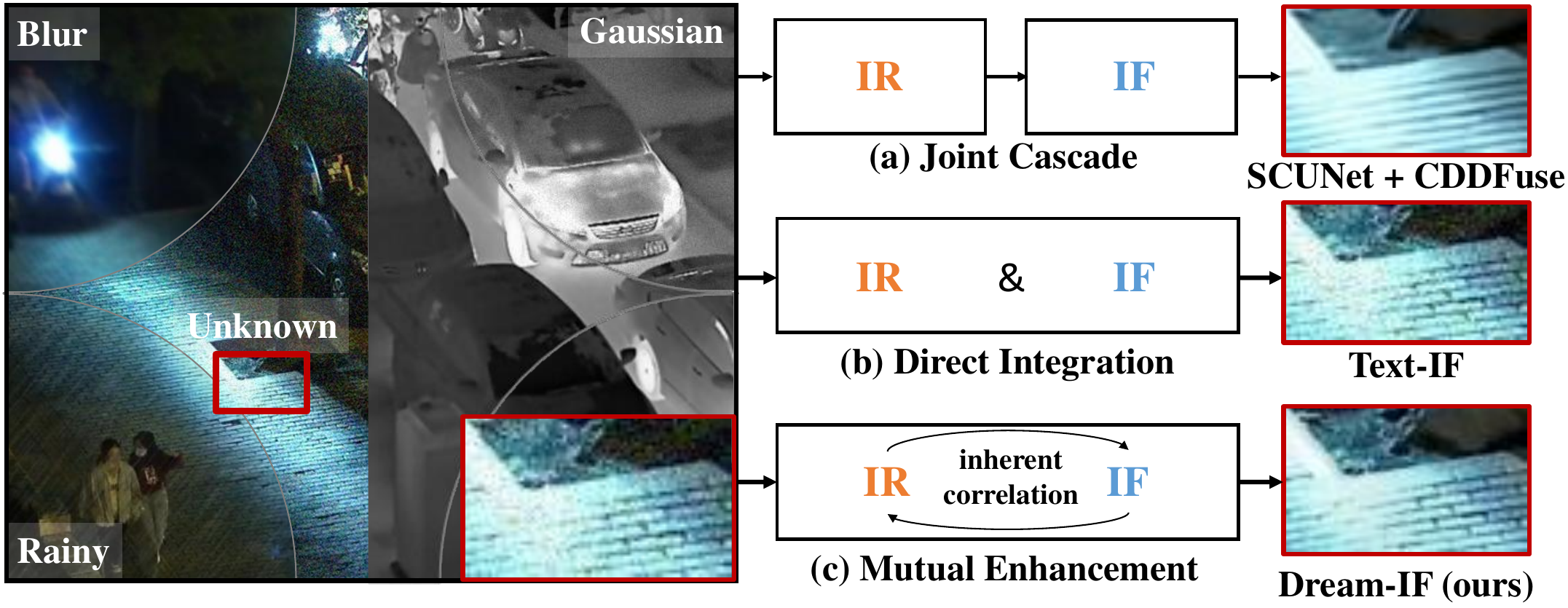}
\caption{
Image fusion (IF) and image restoration (IR) task paradigms. (a) Joint cascaded for restoration followed by fusion, (b) Directly integrate fusion and restoration, (c) Mutually enhanced fusion and restoration through the inherent correlations.
}
\vspace{-13pt}
\label{fig:fig0}
\end{figure}

However, due to sensor malfunctions or environmental disturbances, noise is often present in real-world visible and infrared images~\cite{plotz2017benchmarking,tang2014modeling} resulting in degradation and low quality. Some existing methods~\cite{zamir2022restormer, chen2021hinet} enhance the low-quality images by learning the distribution of datasets and have achieved impressive results. More recently, some approaches~\cite{chen2022simple, wang2022uformer,yang2023multi} effectively handle degradations by capturing long-range pixel interactions through multi-head attention and feed-forward networks, leveraging the relationship to model the restoration process. 

Most existing methods treat image fusion (IF) and image restoration (IR) as two separate tasks~\cite{yang2009multifocus, xia2007novel}, often requiring restoration to be performed prior to fusion, as illustrated in Fig.~\ref{fig:fig0}(a).
Recent studies have attempted to address both tasks within a unified framework, leveraging their shared requirement for effective information extraction.
CU-Net~\cite{deng2020deep} and Deep$\text{M}^2$CDL~\cite{deng2023deepm} designed a unified network to solve general multi-modal image fusion and multi-modal image restoration. It has been demonstrated that multi-modal images can significantly benefit image restoration. 
Text-IF~\cite{yi2024text} and Text-DiFuse~\cite{zhang2025text} directly merge the two tasks without exploring their mutually reinforcing relationship, potentially leading to suboptimal performance, as illustrated in Fig.~\ref{fig:fig0}(b).
Although these works combine image restoration and image fusion in one framework, they still fail to capture their intrinsic coherence.


Intuitively, from the perspective of multi-modal complementarity, image fusion reveals the dominant information of different modalities, which in turn indicates the relative non-dominant information in the other modality, specifically tailoring regions that should be enhanced regardless of degradation issues. By leveraging the relative dominance in image fusion, the difficulty of image degradation recovery can be mitigated. Simultaneously, the use of image degradation tasks can further enhance the adaptability of image fusion to various degradation scenarios. 

Based on this motivation, we propose a Dynamic Relative EnhAnceMent framework for Image Fusion (\texttt{\textbf{Dream-IF}}), jointly boosting the performance of image fusion and enhancement. Unlike most existing methods, Dream-IF first introduces the natural multi-modal complementarity from image fusion to enhance cross-modal image quality, especially for the corresponding non-dominant regions of the respective modality. 
To perform relative dominance-aware enhancement, we designed the relative enhancement (RE) module to dynamically capture the relative dominance of each modality, which is subsequently employed to facilitate cross-modal enhancement of the relatively weak-quality regions. Consequently, we can obtain the enhanced multi-modal features.
By revealing the comprehensive nature of image fusion and converting the dominant regions of one modality to guide the enhancement of the other modality, we enhance the model’s capacity for image fusion in complex conditions, making the fusion model more robust to degradations.
To the best of our knowledge, this work is the first to explore the intrinsic correlation of image enhancement and fusion using comprehensive relative dominance, rather than treating them as separate data-driven tasks or merely integrating them into a unified framework. Our model copes with robust image fusion for image degradation and even broader image enhancement.
The main contributions can be summarized as follows:

\begin{itemize}
\item We for the first time investigate the complementarity of image fusion from the perspective of image enhancement. Based on this, we propose a Dynamic Relative EnhAnceMent framework for Image Fusion, termed \texttt{\textbf{Dream-IF}}, qualified to perform robust image fusion for image degradation and even broader enhancement. 

\item We design a relative enhancement module, which includes a cross enhancement (CE) module that captures the dynamic relative dominance of each modality to specifically enhance the deficient regions in the other modality, and a self enhancement (SE) module, which further restores the image dynamically using prompts guided by the relative dominance.

\item  We conduct extensive experiments to evaluate the effectiveness of our framework. Experimental results demonstrate impressive results and generalization, surpassing other competitive methods in both subjective assessments and objective comparisons.
\end{itemize}
\begin{figure*}
\centering
\includegraphics[width=1\linewidth]{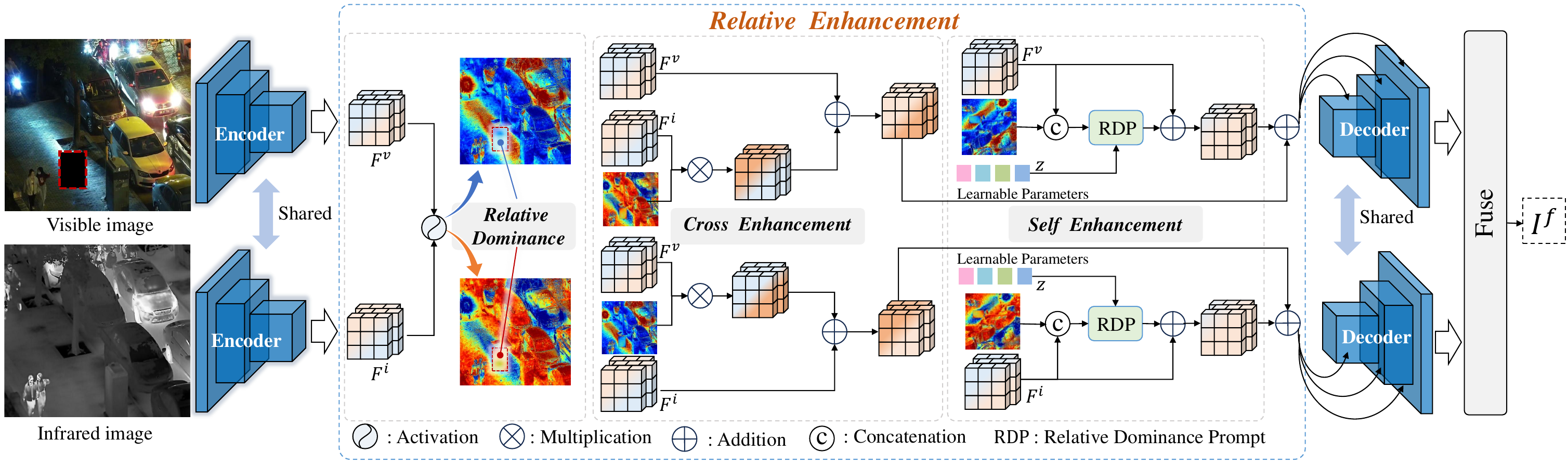}
\caption{An overview of the proposed Dream-IF. We introduce the Relative Enhancement block, which implicitly enhances non-dominant representations by leveraging relative information during the fusion and restoration process. This module captures the relative dominance inherent in the complementary nature of the image fusion model and uses it to facilitate both cross and self enhancement, ultimately producing the enhanced feature.}
\label{fig:framework}
\end{figure*}

\section{Related work}
\label{sec:related}

\noindent\textbf{Image Fusion.} 
Image fusion focuses on extracting effective information from multi-source images and producing an image containing complementary information. 
Traditional methods often perform image fusion by mathematical transformations, with multiscale analysis~\cite{zhang1999categorization}, sparse representation~\cite{zhang2018sparse}, subspace learning~\cite{fu2008multiple}, etc.
With the development of deep learning, CNN-based methods have made significant advancements. DenseFuse~\cite{li2018densefuse}
pioneered the use of deep learning models for fusing infrared and visible images. Inspired by transform-domain image fusion, IFCNN~\cite{zhang2020ifcnn} introduced a general image fusion framework for various fusion tasks. Furthermore, U2Fusion~\cite{xu2020u2fusion} addresses various fusion tasks through a unified, unsupervised image fusion network. Similar fusion tasks with aligned objectives can facilitate the integration of complementary information through cross-task commonality. More recently, TC-MoA~\cite{zhu2024task} dynamically selects task-customized mixture of adapters to capture task generality while preserving unique task characteristics. TTD~\cite{cao2025test} proposed an effective test-time fusion paradigm based on generalization theory proving dynamic fusion is superior to static fusion.
Considering the potential low-quality of different sensors, some researchers~\cite{deng2020deep, deng2023deepm} have begun to recognize the correlation
between image restoration and fusion. DIVFusion~\cite{tang2023divfusion} enhances low-light images to produce a daytime-like fused image with improved visual perception.
Different from these techniques, we first attempt to introduce the naturally complementary dominance of image fusion to dynamically enhance the image quality of non-dominant modality, improving the overall fusion results.


\noindent\textbf{Image Restoration.}
In recent years, 
image restoration techniques~\cite{liang2021swinir, zamir2022restormer, chen2022simple, wang2022uformer} have achieved significant progress.
Particularly, blind restoration~\cite{wu2022blind, huang2020joint, soh2022variational} has gained considerable attention. 
Blind degradation problem is inverse where the degradation is not explicitly known, i.e.,
non-blind methods can model degradation using prior knowledge (e.g., a predefined blur convolution kernel)  while blind methods only assume knowledge of the type of degradation (e.g., blurring)\cite{chihaoui2024blind}. 
ECycleGAN ~\cite{wu2022blind} tackled the blind reserve problem as an image-to-image task while preserving the fidelity of the reconstructed image. BlindDPS~\cite{chung2023parallel} solved blind inverse problems by constructing a separate diffusion prior for the forward operator. AirNet~\cite{li2022all} leveraged consistency to learn the degradation representation, recovering various degraded conditions. PromptIR~\cite{potlapalli2024promptir} introduces a prompt-based learning approach to encode degradation-specific features with prompts and dynamically guide all-in-one image restoration. Despite the extensive exploration of image restoration, these approaches fail to consider the inherent correlation between image restoration and fusion, specifically using the complementarity in image fusion to assist restoration.

\section{Method}    
\label{sec:method}
We present the pipeline of our Dream-IF in Fig.~\ref{fig:framework}, a dynamic relative enhancement framework acting on the fusion model, which leverages relative dominant prompts to perform cross-modal enhancement for non-dominant regions, thereby facilitating a more robust fusion. 

\subsection{The Overview of Dream-IF } \label{sec:overview}

Dream-IF consists of encoders and decoders containing Restormer blocks~\cite{zamir2022restormer} corresponding to different modalities, which can be defined as $\mathcal{E}$ and $\mathcal{D}$ represent the encoder and decoder, respectively. Given the visible image $I^{v}\in\mathbb{R}^{H\times\ W\times3}$ and the infrared image $I^{i}\in\mathbb{R}^{H\times\ W\times3}$, the feature of each modality can be obtained as follows:
\begin{align}
            F^m &=\mathcal{E}(I^m), \\
            F_i^m &= \mathcal{D}_i(F_{i-1}^m),\\
            I^f &= \mathcal{F}(\sum\nolimits_{k\in m}{F^k_N})
\end{align}
where $F^m \in \mathbb{R}^{h\times w} (m\in\{i,v\})$ represents the latent feature of the given modality, the $F^m_i$ denote the feature at the $i$th and $N$ is the number of decoder layer. Our goal is to fuse the complementary information from the two source images and obtain a fused image $I^f$. Thus, the fusion module can be formulated as $\mathcal{F}$, which consists of a Restormer block and a convolution layer to reconstruct the fused image from features.

\subsection{Relative Enhancement} \label{sec:ProCE}

The encoder and decoder tend to extract contemporary information from source images, which means that when a region of one modality is advantageous, another modality is probably disadvantageous. Inspired by the comprehensive nature of image fusion, we proposed a relative enhancement module to enhance the deficient regions of one modality by capturing the relative dominance of another modality, thus generating a more robust image enhancement and fusion.

\noindent{\textbf{Relative Dominance.}}
As discussed above, the image fusion process inherently aims to exploit complementary information from multiple sources. To further enhance this capability, we introduce a mechanism that dynamically preserves complementary features while suppressing redundant information. Specifically, we define the fusion weight for each source feature as its Relative Dominance (RD), formulated as $\text{RD}^m= \sigma \text{Conv}(F^m)$, where $\sigma$ denotes an activation function. The optimization objective is defined as:
 \begin{equation}
      \min\mathbb{E}[\log(1-\sum_{h,w}^N\text{RD}^{(m)})]. 
 \end{equation}
The fusion process aims at capturing the dominance of each source and leveraging it to enhance non-dominant areas. The visualizations of RD are shown in the Appendix.

\begin{figure*}
\centering
\includegraphics[width=1\linewidth]{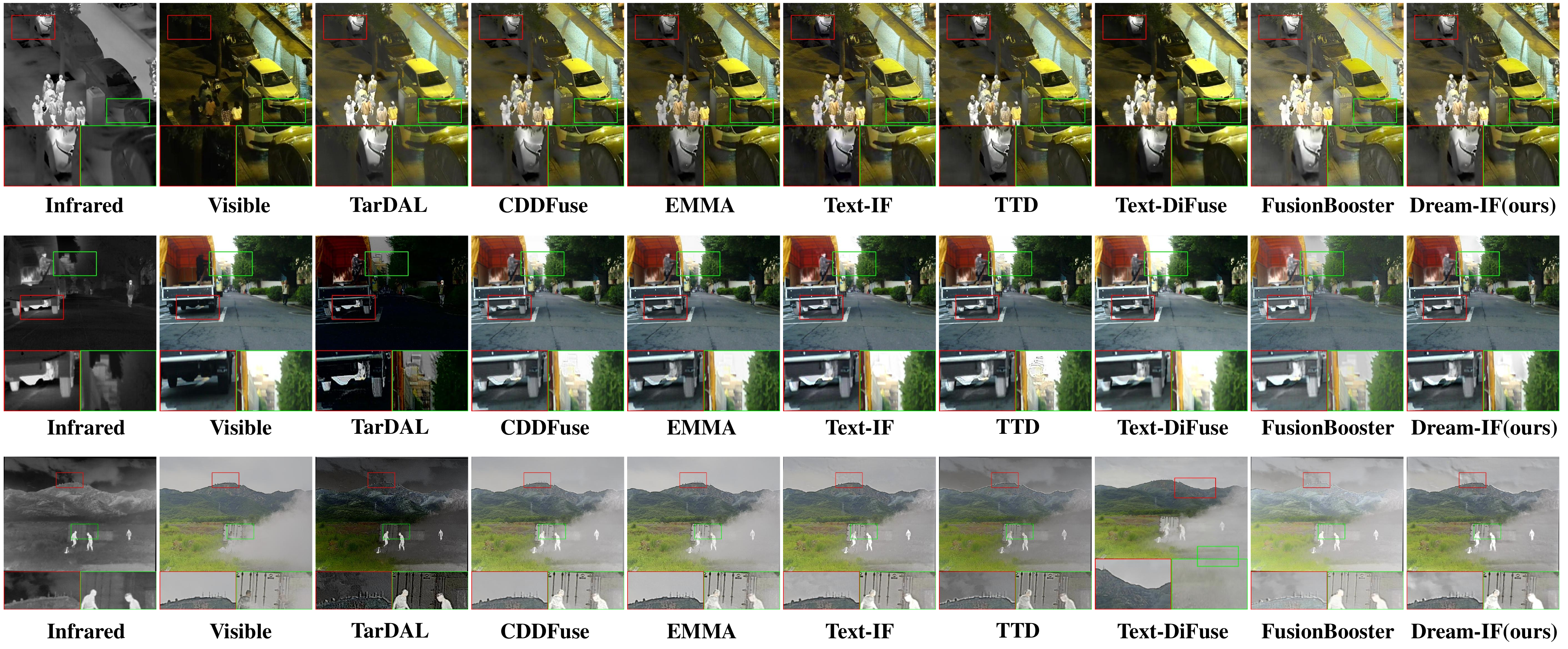}
\caption{Qualitative comparisons of various methods on representative images selected from the LLVIP, MSRS and M3FD datasets.}
\label{fig:compare}
\end{figure*}
\noindent{\textbf{Relative Enhancement.}}
In image fusion, the dominance of one modality can be approximated by the non-dominance of another modality, which should be appropriately enhanced. Thus, the primary goal of the relative enhancement (RE) module is to boost the representation of non-dominant features by facilitating the relative dominance of another modality.

To naturally integrate the RD and the feature to be strengthened, we present the cross enhancement (CE) and self enhancement (SE) module to enhance. The RD denotes the relative dominance from the other modality and indicates its deficient parts to be enhanced, thereby efficiently guiding the reconstruction process.
Specifically, the CE module addresses deficiencies in one modality by complementing it with information from the other modality. It leverages the complementary nature of the modalities, filling in the gaps of the defective feature with the more dominant or reliable feature from the other modality. This ensures a more complete and accurate representation of the scene. The restoration process is guided by the dominant region in the other modality, where the strength of one modality helps identify areas in the other that require enhancement. By focusing on these dominant regions, the module effectively restores the missing or degraded information in the subordinate modality, resulting in a more coherent and refined output. It can be formulated as 
\begin{equation}
    \dot{F}^m=F^m+\textbf{CE}(F^m,RD).
\end{equation}
Further to enhance itself and restore degeneration, to make our model fundamentally resistant to degradation~\cite{potlapalli2024promptir}, we generate a Self Enhancement (SE). This is calculated as:
\begin{equation}
   \hat{F}^m= F^m + \textbf{SE}(\textbf{RDP}(F^m,RD, z), F^m)
\end{equation}
where $z$ is a set of learnable parameters. The Relative Dominance Prompt (RDP) serves as an adaptive enhancement module, which generates restoration prompts dynamically based on the input feature $F^m$ and its corresponding relative dominance RD. Further architectural details are provided in the Appendix.
Afterward, the refined feature is sent to the next block in the decoder, where it is transformed into the denoised and restored feature:
\begin{equation}
 F_{l}^m =\mathcal{D}_l(\widetilde{F}_{l-1}^m),
\end{equation}
where $\widetilde{F}=\text{Conv}(\textbf{Cat}[\dot{F};\hat{F}])$. This enhanced feature is then passed to the fusion block, which combines the features from the two modalities and generates the final fused image.

\subsection{Loss Function} \label{sec:loss}
Dream-IF trains with Gaussian noise degradation, add in clean data. The loss functions we used, including pixel loss, gradient loss, SSIM loss, and color loss, which can be depicted as:
\begin{equation}
    \mathcal{L}_{f} = \mathcal{L}_{pixel} + \mathcal{L}_{grad} + \mathcal{L}_{SSIM}  + \mathcal{L}_{color}.
\end{equation}
Moreover, to explicitly optimize the relative dominance (RD), we define the following loss function:
\begin{equation}
    \mathcal{L}_{RD} = \sum\nolimits_{j}^h\sum\nolimits_{k}^w\sum\nolimits_{m}^NRD^m_{i,j}-1.
\end{equation}
The overall loss function is defined as $\mathcal{L}=\mathcal{L}_f+\mathcal{L}_{\text{RD}}$.

\section{Experiments}
\label{sec:experiments}
In this section, 
We conduct comprehensive qualitative and quantitative comparisons to evaluate the performance of our proposed method against state-of-the-art approaches.
To further validate the contribution of each component, we perform an extensive ablation study.
\subsection{Experimental Setting}
\noindent\textbf{Implementation Details.} 
Our experimental framework is implemented using PyTorch~\cite{paszke2019pytorch} and executed on an NVIDIA 3090 GPU. The proposed architecture adopts a U-Net-based design with a symmetric 4-level encoder-decoder structure, where the encoder extracts hierarchical features and the decoder reconstructs the output image. Each level of the encoder and decoder incorporates multiple Restormer~\cite{zamir2022restormer} blocks, with the number of blocks progressively increasing from higher to lower levels. The complete architectural details, including layer configurations and hyperparameters, are provided in the supplementary material for reproducibility.

\begin{table*}
\centering
\footnotesize

\setlength{\tabcolsep}{0.9mm}
\begin{tabular}{l|ccccc|ccccc|ccccc}
\toprule
\multirow{2}{*}{Method}
& \multicolumn{5}{c}{\textbf{LLVIP Dataset}} & \multicolumn{5}{|c}{\textbf{MSRS Dataset}}  & \multicolumn{5}{|c}{\textbf{M3FD Dataset}}\\ 
& AG & SF &  $Q^{abf}$ & SSIM & VIFF & AG & SF &  $Q^{abf}$ & SSIM & VIFF & AG & SF &  $Q^{abf}$ & SSIM & VIFF  \\
\midrule 

TarDAL & $4.921$ & $18.207$ & $0.410$ & $1.080$ & $0.537$ 
& $1.914$ & $5.944$	 & $0.170$ & $0.930$ & $0.380 $
 & $4.331$ & $15.766$ & $0.292$ & $0.731$ & $0.486$

\\
CDDFuse & $5.403$ & $18.495$ & $0.582$ & $1.184$ & $0.660$
 & $3.779$ & $11.570$ & $0.585$ & $1.303$ & $0.717$
 & $4.800$ & $14.709$ & $0.521$ & $1.379$ & $0.578$
\\

 EMMA
 & $5.560$ & $17.190$ & $0.552$ & $1.192$ & $0.648$ 
& $3.775$ & $11.559$ & $0.621$ & $1.324$ & $0.738$ 
& $\mathbf{5.362}$ & $15.304$ & $0.590$ & $\underline{1.377}$ & $\mathbf{0.711}$ 
 \\ 
 Text-IF 
 & $\underline{5.682}$ & $17.718$ & $\underline{0.640}$ & $1.074$ & $\mathbf{0.813}$ 
 & $3.801$ & $\underline{11.868}$ & $\underline{0.624}$ & $1.117$ & $\mathbf{0.867}$ 
 & $5.034$ & $15.494$ & $\mathbf{0.644}$ & $1.339$ & $0.637$ 
 \\
TTD 
 & $5.509$ & $\underline{19.914}$ & $0.627$ & $\underline{1.196}$ & $0.730$
 & $3.705$ & $11.527$ & $0.549$ & $\underline{1.349}$ & $0.656$
 & $5.022$ & $15.254$ & $0.519$ & $1.367$ & $0.607$

\\
Text-Dif 
 & $4.848$ & $15.528$ & $0.403$ & $1.045$ & $0.524$
 & $\underline{3.842}$ & $11.506$ & $0.436$ & $0.935$ & $0.703$
 & $2.810$ & $8.465$ & $0.149$ & $1.122$ & $0.165$

\\
FusionBooster
 & $5.476$ & $15.464$ & $0.407$ & $0.978$ & $0.551$
 & $3.208$ & $9.012$ & $0.420$ & $1.043$ & $0.630$
 & $3.559$ & $10.245$ & $0.396$ & $1.297$ & $0.515$

\\
 Dream-IF
 & $\mathbf{5.926}$ & $\mathbf{19.992}$ & $\mathbf{0.679}$ & $\mathbf{1.198}$ & $\underline{0.800}$
 & $\mathbf{3.962}$ & $\mathbf{12.293}$ & $\mathbf{0.631}$ & $\mathbf{1.362}$ & $\underline{0.849}$
 & $\underline{5.124}$ & $\mathbf{16.198}$ & $\underline{0.614}$ & $\mathbf{1.387}$ & $\underline{0.704}$

 \\
\bottomrule
\end{tabular}
\caption{Quantitative comparison with SOTAs without degradation. The \textbf{bold}/\underline{underline} indicates the best and runner-up.}
\label{tab:comp_metrics}
\end{table*}

\noindent\textbf{Datasets.} 
Our experiments are conducted on three widely recognized publicly available datasets: LLVIP, MSRS, and M3FD. Further details about the datasets are provided in the supplement.
Following~\cite{zhu2024task}, utilizing $12,025$ image pairs from the LLVIP dataset with random degradation for training and $70$ image pairs for testing.
To rigorously assess the generalization capability of the proposed method, we tested the model on both the MSRS and M3FD datasets without any fine-tuning.

\noindent\textbf{Competing Methods.} 
We compared our approach with seven recent competing methods, including TarDAL~\cite{liu2022target}, CDDFuse~\cite{zhao2023cddfuse}, EMMA~\cite{zhao2024equivariant},  Text-IF~\cite{yi2024text}, TTD~\cite{cao2025test}, Text-DiFuse~\cite{zhang2025text}, and FusionBooster~\cite{cheng2025fusionbooster}.

\noindent\textbf{Evaluation Metrics.} We evaluated the fusion results quantitatively on five metrics, including the average gradient (AG), spatial frequency (SF), gradient-based similarity measurement $Q^{abf}$~\cite{piella2003new}, structural similarity (SSIM)~\cite{wang2004image}, and visual information fidelity for fusion (VIFF)~\cite{han2013new}.

\subsection{Comparison without Degradation}
We conduct comprehensive qualitative and quantitative evaluations to compare the performance of the proposed Dream-IF against state-of-the-art competing methods under the scenario without degradation.

\noindent\textit{Qualitative Comparison.} 
Our comprehensive evaluation reveals that, while all competing fusion methods are capable of combining primary features from infrared and visible images to some extent, the proposed method demonstrates clear advantages that significantly enhance fusion quality.
Our method excels in emphasizing the critical characteristics of infrared images while preserving complementary visible information. As illustrated in Fig.~\ref{fig:compare} (LLVIP dataset, red boxes), the visible images in low-light conditions are often blurred, whereas the infrared images provide essential complementary details. Competing methods, however, suffer from significant infrared content loss in the fusion results, as evidenced by the dark and indistinct details of the cars in the green boxes. Although CDDFuse shows competitive visual performance in the red boxes, it fails to retain sufficient infrared information, leading to suboptimal fusion in the green boxes. In contrast, our method enhances vulnerable modalities through a robust feature integration mechanism, effectively preserving both infrared and visible information. This results in superior perceptual quality and improved information retention, as demonstrated by the clear and detailed fusion results in both highlighted regions. The robust experimental results show consistent performance on both the MSRS and M3FD datasets.


\noindent\textit{Quantitative Comparison.}
The quantitative evaluation, conducted using five widely recognized metrics across the LLVIP, MSRS, and M3FD datasets, is summarized in Table~\ref{tab:comp_metrics}.
On the LLVIP dataset, our method achieves the highest scores across most metrics, namely AG, SF, $Q^{abf}$, and SSIM. Notably, the highest AG and SF scores indicate that our method retains the richest information and sharpest details, attributed to its ability to enhance non-dominant regions while preserving dominant features dynamically. The superior $Q_{abg}$ score reflects better alignment of local gradients and intensities between the source and fused images, demonstrating our method's capability to effectively integrate multi-modal information. Furthermore, the highest SSIM score confirms that the fused images retain a substantial amount of visual information from the source modalities, underscoring the high perceptual quality of our results. These quantitative findings align with the qualitative observations, validating that our method achieves superior fusion performance through dynamic relative enhancement and robust multi-modal integration. Our method also demonstrates competitive performance across most evaluation metrics on the MSRS and M3FD datasets.
The consistency across datasets further validates the generalizability and robustness of our approach under diverse scenarios.

Overall, the quantitative results reinforce the effectiveness of our method in achieving high-quality image fusion, as evidenced by its ability to retain rich information, preserve structural details, and align multi-modal features effectively.

\begin{table}
\centering

\setlength{\tabcolsep}{1.0mm}

\begin{tabular}{l|ccccc}

\toprule
Methods & AG & SF & $Q^{abf}$ & SSIM & VIFF\\
\midrule 

TarDAL & $4.745	$ & $17.337$ & $0.266$ & $	0.503$ & $0.256	$
\\
CDDFuse &$4.530$ & $15.929$ & $0.523$ & $1.179$ & $0.618$
 \\
EMMA & $4.593$ & $14.818$ & $0.499$ & $1.186$ & $0.592	$
 \\

Text-IF &	$4.863$ & $16.554$ & $0.536$ & $1.156$ & $0.609	$
\\

TTD&$4.948$ & $16.637$ & $0.489	$ & $1.144$ & $0.533$

\\
Text-DiFuse& $4.084$ & $13.777$ & $0.364$ & $1.049$ & $0.490$
\\
FusionBooster& $4.524$ & $13.401$ & $0.380$ & $1.017$ & $0.533$
\\
Dream-IF & $\mathbf{5.243}$ & $\mathbf{17.441}$ & $\mathbf{0.580}$ & $\mathbf{1.186}$ & $\mathbf{0.705}$
\\
\bottomrule
\end{tabular}
\caption{Comparison with SOTAs under degradation. The \textbf{bold} indicates the best.}
\label{tab:compare_degra}
\end{table}

\subsection{Comparison with Degradation}

To assess the robustness and real-world applicability of our method, we introduce a diverse set of degradations~\cite{zhang2021designing} to simulate challenging scenarios that amplify modality discrepancies. These degradations include Gaussian noise, Poisson noise, speckle noise, and other common artifacts encountered in practical applications (e.g., blur, etc.). Additionally, we incorporate synthetic rainy degradations to further mimic real-world conditions. For a fair comparison, we compare existing methods for image fusion and restoration. This includes two-stage fusion, which performs degradation recovery followed by fusion, and one-stage fusion with degradation.
 
\begin{figure*}
\centering
\includegraphics[width=1\linewidth]{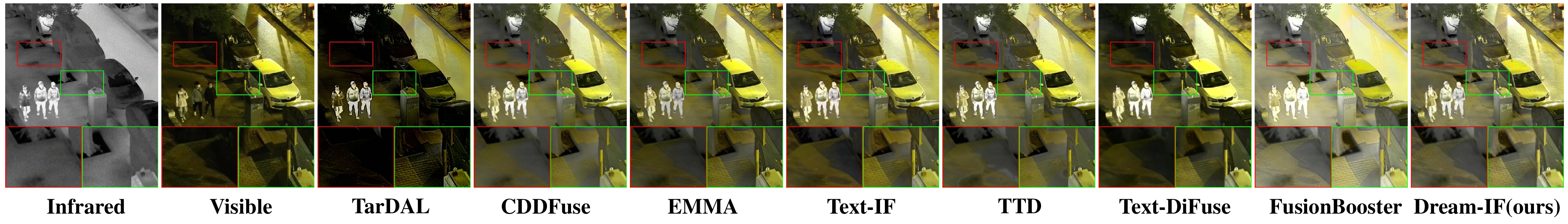}

\caption{Qualitative comparisons of various methods under degradation on the LLVIP dataset. SCUNet+ refers to the application of the SCUNet model for image restoration prior to the fusion process. }

\label{fig:compare_degra}
\end{figure*}


\noindent\textbf{Degradations.} 
Due to the inherent limitations of imaging conditions, various types of noise inevitably arise in practical scenarios, significantly impacting the quality and usability of captured images.
To accurately simulate real-world degradation, we model three fundamental types of noise based on their physical origins in optical imaging systems.

Gaussian Noise primarily results from inherent electronic noise in image acquisition or transmission devices, as well as signal interference in low signal-to-noise ratio (SNR) environments. Mathematically, Gaussian noise can be expressed as:
$y=x+\mathcal{N}(0,\sigma).$
Poisson noise, also referred to as shot noise, arises from the inherent quantum uncertainty in photon counting during the light detection process, which can be formulated as:
$    y=x + \frac{\lambda ^ke^{- \lambda}}{k!}.$
Speckle noise is caused by random phase interference in coherent imaging processes, representing structural corruption rather than additive distortion. It is modeled as:
$    y=x\odot(1+\mathcal{N}(0,\epsilon).$
In this paper, we set $\sigma=35$, $\lambda\in[2,4]$, and $\epsilon\in[2,25]$.

Each training sample undergoes one or more random degradations selected from the aforementioned types. 
Current image fusion methods often struggle with blind degradation, frequently leading to suboptimal fusion results. Our approach addresses this limitation by narrowing the domain gap and applying cross-modal enhancement for restoration to improve the final output. 

\noindent\textbf{Two-stage Fusion.} For blind restoration, we apply random degradation to the LLVIP test set. To ensure a fair comparison, we employ a two-stage fusion strategy. Specifically, for all the competing methods, we adopt SCUNet~\cite{zhang2023practical}, a widely recognized restoration model, to restore images before fusion. For comparison, our Dream-IF directly performs image restoration and fusion in a unified model, which is a more challenging.

\noindent\textit{Quantitative results.} Our method compared with the combined results of SOTA image fusion methods after SCUNet restoration on degraded source images are presented in Fig.~\ref{fig:compare_degra}. Our method retains more details, exhibiting a more realistic overall texture, particularly in the green box. CDDFuse exhibits competitive performance but also suffers from detail and texture loss. TarDAL suffers from fidelity loss, while Text-IF, TTD, Text-DiFuse and FusionBooster produce excessively smooth results, omitting important information. Although EMMA performs competitively in the green box, it fails to preserve details in the red box as effectively as ours. The advantages demonstrate that our approach effectively performs both restoration and fusion, with its strengths becoming more pronounced as the differences between modalities increase.

\noindent\textit{Qualitative results} are reported in Table~\ref{tab:compare_degra}, demonstrating that our method achieves competitive results even without the use of a specialized restoration model. Specifically, our method achieves optimal values across all metrics. The highest AG and SF scores indicate that Dream-IF excels at retaining texture and detail. VIFF, SSIM, and $Q^{abf}$ indicate that our method effectively preserves mutual information between the source images, ensuring details and contextual information are retained in the output while maintaining subjective visual quality. It underscores the ability of our approach to preserve image fidelity and enhance degraded features without relying on a dedicated restoration model.

\noindent\textbf{One-stage Fusion.}
To verify the performance of our method in simultaneous fusion and restoration within a single framework, we further evaluate our Dream-IF with Text-IF, a text-guided restoration and fusion model. Since Text-IF requires descriptive text to specify the degradation type, we provide degradation text guidance for Text-IF.
It is worth noting that our model does not take any prior knowledge of the degradation type, which is a much more challenging setting to handle blind restoration and fusion.
We evaluated image restoration using the learned perceptual image patch similarity (LPIPS). LPIPS captures perceptual differences between images, with lower scores indicating greater similarity.
\begin{table}
\centering
\setlength{\tabcolsep}{0.3mm}

\begin{tabular}{l|ccccccc}
\toprule
 Methods & Rain  & Gaussian & Poisson  & Speckle  & Blur & Resize \\
\midrule 
Text-IF*& $0.273$ & $0.275$ & $0.275$ & $0.270$ & $0.269$  & $0.244$\\
Text-DiFuse & $0.296$ & $0.291$ & $0.268$ & $0.266$ & $0.290$ & $0.268$ \\
Dream-IF & $\mathbf{0.241}$ & $\mathbf{0.255}$ & $\mathbf{0.242}$ & $\mathbf{0.240}$ & $ \mathbf{0.260}$ & $\mathbf{0.241}$\\

\bottomrule
\end{tabular}
\caption{One-stage fusion comparison. \textbf{Bold} is the best. Text-IF* is Text-IF with description text.}
\label{tab:degra}
\end{table}

\begin{figure}
\centering
\includegraphics[width=1\linewidth]{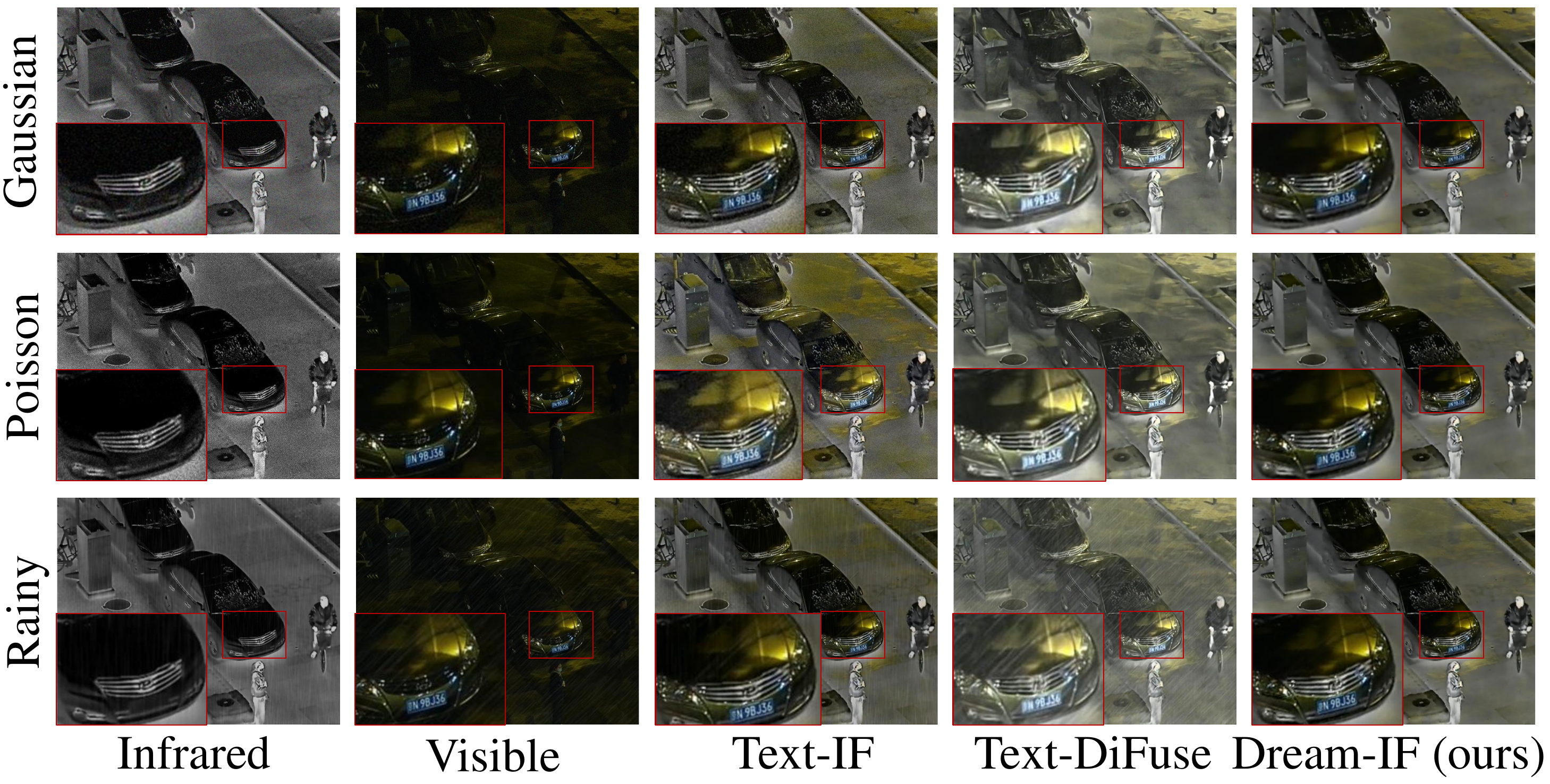}

\caption{Visualization of comparison in one-stage restoration.}

\label{fig:compare_degra_textif}
\end{figure}

\noindent\textit{Qualitative results.} As presented in Fig.~\ref{fig:compare_degra_textif}, it demonstrates that our method recovers more texture details affected by degradation. However, even with degradation descriptions, Text-IF fails to restore images affected by Gaussian noise fully, and it results in overly smooth fused images when applied to other types of noise. Text-DiFuse removes noise through diffusion, but yields poor performance. This demonstrates that our method is robust to various degradations, benefiting from dynamic dominance enhancement to improve inferior features. In contrast, our method leverages information from both modalities to seamlessly integrate image fusion and restoration, enabling effective blind restoration and fusion. It further validates our effectiveness in utilizing the comprehensive nature of fusion to drive overall image enhancement in the multi-modal fusion task.

\noindent\textit{Quantitative results}. As shown in Table~\ref{tab:degra}, our method consistently outperforms in terms of LPIPS scores across most degradation types. Notably, the method performs exceptionally well when applied to images affected by Gaussian noise. Gaussian noise is particularly challenging because it introduces subtle distortions that affect all frequencies within the image. These distortions are difficult to mitigate without compromising finer image details or introducing artifacts. The lower LPIPS scores we achieve in this context are indicative of higher perceptual similarity, meaning that our approach is better at maintaining the natural textures and structures of the image. This ability to preserve details without significant loss of fidelity means that our framework produces results that are visually closer to the original, real image, making it a highly effective solution for addressing Gaussian noise degradation. Furthermore, the preservation of fine-grained textures and intricate features enhances the overall quality of the images processed by our method, distinguishing it from other approaches that might smooth or blur these details.

\begin{table}
\centering

\setlength{\tabcolsep}{0.9mm}

\begin{tabular}{cccc|ccccc}
\toprule
Base & RD & CE & SE & EI  & AG & PSNR &  $Q^{abf}$ & VIFF \\
\midrule     
 $\checkmark$ & -& -&  -& $5.402$ & $18.032$ & $0.627$ & $1.181$ & $0.736$

\\
 $\checkmark$ & $\checkmark$ & -& - & $5.638$ & $18.594$ & $0.659$ & $1.185$ & $0.756$
\\
$\checkmark$ &$\checkmark$ & $\checkmark$ & -& $ 5.642$ & $18.633$ & $0.675$ & $1.193$ & $0.770
$ 
\\

$\checkmark$&$\checkmark$  &- &$\checkmark$ & $5.746$ & $19.446$ & $0.668$ & $1.191$ & $0.797$ \\

$\checkmark$& $\checkmark$ & $\checkmark$ & $\checkmark$ & $\mathbf{5.926}$ & $\mathbf{19.992}$ & $\mathbf{0.679}$ & $\mathbf{1.198}$ & $\mathbf{0.800}$
 \\
\bottomrule
\end{tabular}

\caption{Ablation studies on LLVIP dataset. \textbf{Bold} is the best.}
\label{tab:ablation}
\end{table}

\begin{table}
\centering

\setlength{\tabcolsep}{2.4mm}

\begin{tabular}{l|ccc}
\toprule
Methods  & Recall & mAP@.5  & mAP@.5:.95
\\
\midrule 
TarDAL  & $0.8850$ & $0.9495$ & $0.6291$
\\
CDDFuse   & $0.8927$ & $0.9496$ & $0.6031$ 
\\
EMMA  & $0.8894$ & $0.9502$ & $0.6192 $\\
Text-IF & $0.8951$ & $0.9517$ & $0.6148$\\
TTD & $0.8888$ & $0.9497$ & $0.6252$\\
Text-DiFuse& $0.8742$ & $0.9394$ & $0.5964$\\
FusionBooster& $0.8925$ & $0.9502$ & $0.6245$ \\
Dream-IF & $\mathbf{0.9041}$ & $\mathbf{0.9543}$ & $\mathbf{0.6327}$\\
\bottomrule
\end{tabular}

\caption{Comparison of object detection on LLVIP dataset. \textbf{Blod} indicates the best.}
\label{tab:det}
\end{table}

\subsection{Performance on Downstream Task}
\label{sec:task}
We perform experiments on object detection to verify the compatibility of Dream-IF with high-level downstream tasks. Specifically, we train the YOLOv11~\cite{khanam2024yolov11} detector on the LLVIP dataset, using fusion results generated by competitive comparison methods, and evaluate performance using recall and mAP. As shown in Table~\ref{tab:det}, Dream-IF outperforms the competition in object detection, achieving the best performance in recall, mAP@.5, and mAP.5:.95. Additionally, visualizations of the comparisons in object detection are shown in the Appendix.



\subsection{Ablation Study}
We conduct ablation studies on the LLVIP dataset, as shown in Table~\ref{tab:ablation}, to validate the effectiveness of the proposed CE and SE module. Specifically, we evaluate the performance of our model by removing the CE and SE modules, treating the model without these components as the baseline. The results demonstrate that adding either the CE or SE module leads to minor improvements in the performance across all evaluation metrics. When both modules are incorporated into the model, substantial improvements in all indicators are observed. These results highlight the distinct and complementary contributions of each module. Notably, the inclusion of both CE and SE along with the final model configuration leads to the highest scores across all metrics, confirming that our method achieves the best qualitative and quantitative performance among all ablation settings. This further validates the importance of the CE and SE modules, as well as their combined effect in enhancing the overall model's capabilities. The improvements across various indicators demonstrate the efficacy of our approach in achieving superior results.


\section{Conclusion}
\label{sec:conclusion}
In this paper, we first explore the inherent connection between image fusion and image enhancement. Based on the observation that the dominant region of one modality in image fusion relatively indicates the inferiors of the other modality, we propose a dynamic relative enhancement framework for image fusion (Dream-IF). We extract the relative dominance from the fusion model for each sample and use it to perform prompting cross-modal enhancement. Notably, our Dream-IF is a blind restoration model, coping with robust image fusion for image degradation and even broader image enhancement. Extensive experiments validate our effectiveness against the competing methods. In the future, we will further study the potential of relative dominance derived from image fusion in other related tasks.


\section{Acknowledgments}
\label{sec:acknowledge}
This work was sponsored by the National Natural Science Foundation of China (No.s 62222608, 62436002, 62476198), the Tianjin Natural Science Funds for Distinguished YoungScholar (No. 23JCJQJC00270), the Zhejiang ProvincialNatural Science Foundation of China (No. LD24F020004), and the Natural Science Foundation of Tianjin (No. 25JCYBJC00950).

   \bibliography{aaai2026}

\end{document}